\documentclass{article}

\usepackage{arxiv}

\usepackage[utf8]{inputenc} 
\usepackage[T1]{fontenc}    
\usepackage{hyperref}       
\usepackage{url}            
\usepackage{booktabs}       
\usepackage{amsfonts}       
\usepackage{nicefrac}       
\usepackage{microtype}      
\usepackage{lipsum}
\usepackage{graphicx}
\usepackage[ruled,vlined]{algorithm2e}
\graphicspath{ {./images/} }

\title{SCNN: Swarm Characteristic Neural Network}

\author{
 Nguyen Ha Thanh \\
  Japan Advanced Institute of Science and Technology\\
  \texttt{nguyenhathanh@jaist.ac.jp} \\
   \And
 Nguyen Le Minh \\
  Japan Advanced Institute of Science and Technology\\
  \texttt{nguyenml@jaist.ac.jp} \\
}

\begin{document}
\maketitle
\begin{abstract}
Deep learning is a powerful approach with good performance on many different tasks. However, these models often require massive computational resources. It is a worrying trend that we increasingly need models that work well on more complex problems. In this paper, we propose and verify the effectiveness and efficiency of SCNN, an innovative neural network inspired by the swarm concept. In addition to introducing the relevant theories, our detailed experiments suggest that fewer parameters may perform better than models with more parameters. Besides, our experiments show that SCNN needs less data than traditional models. That could be an essential hint for problems where there is not much data.
\end{abstract}


\section{INTRODUCTION}
Neural networks have currently become one of the remarkable approaches in machine learning. These models can achieve excellent performance in many tasks due to the diversity in their architecture and their ability to automatically extract features. Automatic feature selection is crucial in our big data world because manually choosing the essential features for any data is either impossible or too costly. This ability is essential for neural networks to outperform other AI algorithms and even humans in some tasks. Deep Mind’s Alpha Go is one of the most well-known events in which AI surpasses human ability by the power of the neural network.

Observing the development of artificial neural networks, we can see that the number of parameters in these models is increasing. This trend will obviously exacerbate the information catastrophe \cite{vopson2020information}. In addition, the steady growth of hardware power in line with Moore Law’s predictions will continue to support this trend. In 2018, the BERT \cite{devlin2018bert} model on transformer architecture had 110 million parameters that can perform impressively in a wide range of tasks. Two years later, in 2020, OpenAI released the GPT-3 \cite{brown2020language} model with 175 billion parameters. This trend can be a valid idea supporting the belief that the more complex the model, the better the performance.

Along with the neural network, another effective direction for machine learning is swarm intelligence. Swarm studies appear in many different domains. This approach is much cheaper in terms of complexity than neural networks. The general idea is to use interactions between individuals and the environment to make decisions. Previous studies consider the swarm as an active object that can achieve the goal by distributed intelligence of its member.

Yang et al. \cite{yang2013swarm} indicated different reasons for the developing prominence of algorithms based on swarm intelligence, in particular the versatility and adaptability offered by the algorithms. The adaptability of algorithms to external factors, together with their simplicity and self-learning capacity, are the key highlights displayed. This has increased interest in developing applications in various areas. From the computational perspective, swarm intelligence models provide algorithms for solving distributed optimization calculations.

\begin{figure}[h]
  \centering
  \includegraphics[width=.6\linewidth]{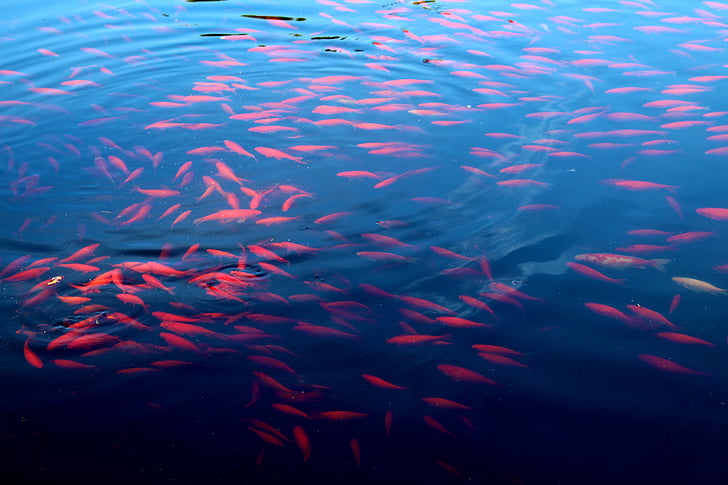}
  \caption{Shoal of koi fish can be identified more easily than a single fish}
  \label{fig:koi}
\end{figure}

In addition to the active aspect, the passive aspect of a swarm is also worth studying. Swarm can be the feature providing essential hints that help to solve various problems. For instance, it seems impossible for humans with healthy eyes to spot and identify a single insect flying miles away. However, this becomes feasible if the insects fly in a swarm. A single isolated individual cannot carry the features of a swarm’s individuals. Figure \ref{fig:koi} shows approximately one shoal of koi fish; they can be identified more easily than a single fish. From this observation, the swarm filter \cite{nguyen2019swarm} was invented as a component in a neural network to extract swarm features automatically. In terms of computation, the model can extract swarm information of an individual as a data feature.

In this paper, we want to show the efficiency of an architecture using a swarm filter, which is the SCNN. We compare the SCNN with the conventional architecture using basic elements of neural networks. We trained all models with a large amount of data from scratch without any pretraining or transfer learning techniques so that we can better verify the learning ability of the networks. In our experiments, the SCNN showed impressive results on both effectiveness and efficiency.

The main contributions of this paper are the introduction of a novel architecture for neural networks using the swarm filter and examining its performance and learning behavior. To convey the paper in a straightforward manner, we cover some background knowledge in Section \ref{sec:background}. The model as well as our detailed experiments and analysis are covered in \ref{sec:scnn}. Section \ref{sec:related_works} distinguishes our work from others. Finally, we present some conclusions and future directions for this model in Section \ref{sec:conclusion}.

\section{BACKGROUND}
\label{sec:background}
\subsection{SWARM INTELLIGENCE}
Swarm intelligence is a field of artificial intelligence that is motivated by the conduct of some social living creatures, for example, ants, termites, birds, and fish. For the most part, decentralized control and self-association are the essential highlights of systems based on swarm intelligence, which result in emergent behavior. Emergent behavior develops through local cooperation among the components of the system, and it cannot be accomplished by any of the individuals of the framework acting alone. Accordingly, swarm intelligence is decentralized, self-arranging and disseminated through a given environment. In the natural habitat of organisms, swarm intelligence systems are usually used to tackle issues such as the relocation of colonies, evasion of prey and foraging for food. Data are stored regularly through participating agents and stored in the surroundings.

Following Dorigo et al. \cite{dorigo2014swarm}, systems that are based on swarm intelligence have the following properties. First, they are made of numerous participants who work collectively with others. In addition, the participants are moderately homogeneous. Dorigo et al. stated that another property is that the connections between the participants depend on straightforward behavioral rules that utilize only the local data that the participants exchange using the environment or directly with others. Likewise, the overall result of the framework comes from the collaborations among the participants in the environment. In other words, the behavior of the group is self-organizing. Accordingly, the characterizing feature of the swarm intelligence framework is its capacity to act in a planned manner without the nearness of an organizer or an outside controller.

Ant colony optimization is one of the most common swarm intelligence algorithms. Lopez-Ibanez et al. \cite{lopez2012automatic} indicated that it is propelled by the social conduct and steering method of ants looking for food. Lim et al. \cite{lim2009innovations} explained that ants can create the shortest path from their colonies to their sources of food and back. While scanning for food, ants initially meander around their environment arbitrarily. After discovering food, they return to their colonies while laying a path of a synthetic substance called a “pheromone” along their way. This pheromone is utilized for communication whereby different ants will identify the trail of the pheromone. Lopez-Ibanez et al. stated that the richer the pheromones along a path, the more probable that different ants will identify and follow the route. Ants will, in general, pick the course set apart by the most potent concentration of pheromone fixation.

The concentration of pheromone vanishes after some time, decreasing its alluring quality. The pheromone will vanish faster if it takes the ants more time to go to the food and return. In this manner, the pheromone along a shorter way will be fortified faster on the grounds that when different ants follow the way, they continue adding their pheromone to the path before it dissipates. As a result, the shortest route between the colonies and the identified sources of food inevitably emerges. From the computational perspective, one of the benefits of the evaporation of the ants’ pheromone is to mitigate convergence to a locally optimal solution. Without a doubt, pheromone dissipation is a valuable mechanism for forcing a type of overlooking, henceforth permitting the solution space to be investigated in an all-inclusive way.

Another conventional swarm intelligence algorithm is known as particle swarm optimization. Garro et al. \cite{garro2015designing} articulated that this swarm intelligence is propelled by social conduct fish schooling as well as the flocking of birds. In this manner, the particle swarm optimization algorithm is a developmental calculation model that has underlying foundations in both natural and artificial life. It uses a populace of particles that fly through a multidimensional solution space with their velocities set \cite{lim2009innovations}. Every particle encodes a solitary convergence of all search measurements. The related position and speed of every particle are arbitrarily created. Algorithm \ref{alg:pso} demonstrates how particle swarm optimization works.

\begin{algorithm}[H]
\label{alg:pso}
\SetAlgoLined

 Initialize each particle and its velocity V and location X\;
 \Repeat{$t$=MAX or $f_{gbest}$=0}{
  \For{each particle i}{
  Calculate the fitness value $f_i$ of particle $i$\;
  \If{$f_i$ \textless $f_{pbest_i}$}{
    $pbest_i$ = $X_i$;\\
  }
  \If{$f_i$ \textless $f_{gbest_i}$}{
    $gbest_i$ = $X_i$;\\
  }
  Update $X_i$ and $V_i$; 
  }
 }
 \caption{Particle Swarm Optimization}
\end{algorithm}

Bee colony optimization is yet another well-known swarm intelligence algorithm. Yuce et al. \cite{yuce2013honey} stated that the bee colony optimization algorithm was motivated by the search practices of honeybees. Foraging honeybees are conveyed from their colonies to scan for quality nourishment sources, for example, blossom patches. After discovering a decent nourishment source, Lim et al. \cite{lim2009innovations} said that a foraging honeybee comes back to the hive and educates its hive mates using waggle dance. The mystery behind the honeybee waggle dance was decoded by Von Frisch \cite{von1967dance}, who identified that the waggle dance is a specialized communication mechanism with different honeybees.

Accordingly, the waggle dance of the foraging honeybee passes on three bits of significant data to different honeybees, that is, the distance to the identified food source, the direction, and the quality of the nourishment source. Specifically, the foraging honeybee utilizes this waggle dance as a way to persuade different honeybees to be followers and return to the nourishment source. Subsequently, more honeybees are pulled into all of the more encouraging nourishment sources. Yuce et al. \cite{yuce2013honey} said that such a system is sent by honeybee colonies to acquire quality nourishment quickly and productively.

\subsection{ARTIFICIAL NEURAL NETWORKS}
These algorithms perceive essential features from data through a procedure that impersonates how the human cerebrum works. A neural network alludes to a system of neurons, which can either be artificial or natural, existing in nature. In an artificial neural network, a neuron is a mathematical function that collects and classifies data. The model is composed of thousands of neurons that are densely interconnected and incrementally calculates the abstraction of the data from the input layer to the output layer. There is a weighted connection between nodes in a neural network. Instead of biology signals as weights for connection, the artificial neural network uses a numerical value. When that value passes a threshold, the neuron fires a new signal to the next one.

There is no single general architecture for every neural network; each of them has a different design to perform best in their application problems. However, all of them are constructed using some basic components. The simplest elements of neural networks are perceptions \cite{rosenblatt1958perceptron}. More complex components are later developed based on them, such as the cells and gates of RNN networks and the convolution units of CNN networks. Each architecture fits specific tasks with its characteristics.

The ﬁrst component in the history of artiﬁcial neural networks is the perceptron. It is a mathematical function that calculates $f(x)=w \cdot x + b$, in which $x$ is the input vector and $w$ and $b$ are weight and bias parameters of the model, respectively. This simple model can only work on linearly separable data.

Overcoming that weakness of perceptron, multilayered perceptron (MLP) was proposed. MLP contains perceptrons that are organized in layers that are inter connected, whereby the information layer gathers the patterns that are input. The layer used for output is composed of output signals as well as classifications to which patterns used during input may outline. Within neural networks, the margin of error is minimized through the hidden layers, which are used for fine-tuning the weightings that are input.
During the training process, the model needs to optimize the parameters so that it can fit with the training data. After being trained, the model can predict the outcome of unseen data in the future.

MLP encounters a challenge of overfitting. Another common neural network that overcame that challenge is known as the convolutional neural network (CNN). A CNN is composed of at least a number of convolutional layers. These layers can either be pooled or interconnected completely. Prior to passing the outcome to the following layer, the convolutional layer utilizes a convolutional function on the information \cite{kim2014convolutional}. Because of this convolutional function, the neural network can be enabled at a much deeper location. Because of this capacity, convolutional neural networks show exceptionally successful results in various applications.

Another common type of neural network is the recurrent neural network (RNN), in which the output of a specific layer is stored and subsequently fed back to the input node. Sundermeyer et al. \cite{sundermeyer2013comparison} stated that this is essential in predicting the results of the layer. Accordingly, following every data item stored and then fed back to the input, every node will recall some data that it had in the previous step. LSTM is an improved neural network based on RNN with a cell architecture that allows for the selection of important information for prediction steps. Every node is used partly as a memory cell while executing operations. The neural system starts with propagation at the front, yet it recalls the data it may require to use far ahead. If the prediction is not correct, then the framework begins the process of self-learning and progresses in the course of making the correct prediction during backpropagation.

\section{SCNN: SWARM CHARACTERISTIC NEURAL NETWORK}
\label{sec:scnn}
\subsection{SWARM FILTER AND SWARM FEATURE}
Swarm filters and swarm features are important concepts proposed by Nguyen et al. \cite{nguyen2019swarm}. These concepts are based on a simple observation of swarming that individuals in a swarm often show distinctive signs from singular individuals or individuals in another swarm. A swarm here is not limited to a natural swarm but rather a symbolic swarm, for example, a company has its own rules about working style, uniformity and communication style that employees must follow. Based on such attributes, we can recognize an individual as an employee of a certain company. As another example, texts are a regular set of words, and based on the features in the word arrangement, we can perceive the meanings to which the texts refer. 

Swarm features are the unique properties of an instance that exists in a swarm. These features may exist explicitly or in latent dimensions. Therefore, they need to be extracted using learning methods such as neural networks. Nguyen et al. proposed the swarm filter as a part of a neural network that extracts these features. In other words, the swarm filter is a set of weights in the neural network. By calculating the outer product with input, it produces swarm features. Swarm filter layers can be stacked and combined with other elements in the neural network.

\subsection{PROPOSED ARCHITECTURE}
\begin{figure}[h]
  \centering
  \includegraphics[width=.8\linewidth]{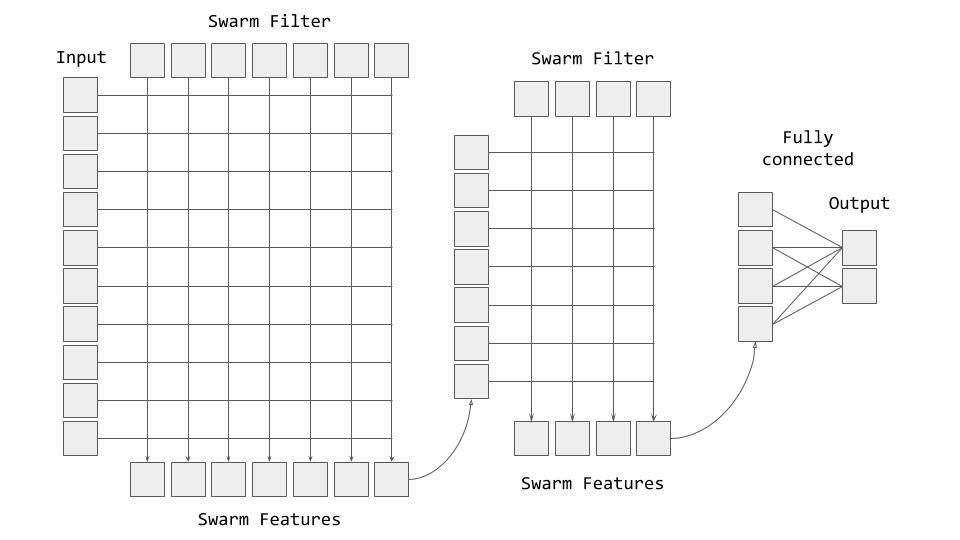}
  \caption{General architecture of SCNN}
  \label{fig:general_architecture}
\end{figure}

The swarm characteristic neural network (SCNN) is a neural network that makes use of the swarm filter \cite{nguyen2019swarm} to extract the features in the latent swarm to which the samples belong. The general architecture of the network is represented in Figure \ref{fig:general_architecture}. This neural network contains two swarm filters and one fully connected layer as its components.

The embedding layer maps the words in text into points in vector space. To determine the model’s learning capability, no pretrained embedding was used. After being trained, the points in the space corresponding to the words will reflect the semantic relationship between them. At its simplest, pairs of words with the same meaning will have a shorter distance than other pairs of words. This is the basis for the model to make a decision on the input sentence based on recognizing the difference in the semantics of words.

The input under the form of a concatenated tensor is fed into the neural network. The purpose of the swarm filter is to make the data reflect their swarm features. Let $x=(x_1,x_2,..,x_n)$ and $s=(s_1,s_2,..,s_m)$ be the data vector and the swarm filter vector, respectively. We can obtain the swarm feature vector $f$ by using Equation \ref{eq:scnn}.

\begin{equation}
\label{eq:scnn}
    f = (f_j | f_j = \frac{1}{n}\sum_{i=1}^{n}{(x \otimes s)_{ij}})\\
\end{equation}

Since the model updates its parameters, the samples sharing the same swarm feature will make the filter value shift in the same direction. Consequently, the model can extract the latent swarm feature after the training process. Swarm filters transform the data to a more abstract level in a smaller dimensional space. In terms of computational efficiency, the swarm filter requires fewer parameters than other standard components in deep learning models. We only need s parameters of a swarm filter to transform the data to s-dimensional space. The fully connected layer converts the information of the final swarm filter layer into meaningful information for the purpose of the model. In other words, with the fully connected layer, the model can mix the signals needed to make the final decision. It can be seen that every element in the SCNN can be fine-tuned through backpropagation. As a result, it can be used as an end-to-end model in learning problems.

SCNN is quite straightforward in its idea. If we consider a single input combined with the weight in the swarm filter as a particle in a latent swarm, they facsimile the behavior that is described in swarm intelligence in general. Before the network is trained, each particle follows a random direction that is constrained by the shared weights in the swarm filter. The loss function assesses how good the random solutions are. By the process of backpropagation, every particle must follow the best solution with the minimized loss. The swarm filter in this context can be considered the mean for particles to communicate and share information about the reward. This observation complies with the statement of Dorigo et al. \cite{dorigo2014swarm} about particle communication.

\subsection{EXPERIMENTS AND ANALYSIS}
\label{sec:exp}
In this section, we conduct different experiments to learn how the proposed architecture works. First, we study the performance of the model and examine the model’s effectiveness and efficiency. After that, we analyze the learning curve on the experimental data of the model compared to other basic models in deep learning. Finally, the swarm feature visualization in the last experiment provides an intuitive understanding of how the data are processed in the information flow of the model.

\subsubsection{Task and Dataset}
\label{sec:task_and_data}
We run the experiment on a sentiment analysis task using SCNN and other deep learning architectures to compare their performance. Sentiment analysis is a tool in social media that focuses on the qualitative aspect of views, likes and retweets about a particular product. It came about through the eagerness of people to express and share their opinions. Sentiment analysis is an automated process that analyses text data and sorts them into positive, negative or neutral comments \cite{saif2012semantic}. Given a sentence, the model needs to predict whether its sentiment is positive or negative.

To verify the efficiency in terms of calculation, we compare the number of parameters of the models. We also analyze in detail the values of the swarm feature tensors attained from the trained model. From the point of view of deep learning, the better features the model can extract from the data, the better performance the model can achieve. The dataset for the experiment is Sentiment 140 \cite{go2009twitter}.

\begin{figure}[h]
  \centering
  \includegraphics[width=.6\linewidth]{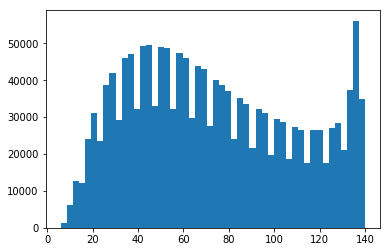}
  \caption{Distribution of sample length in training set.}
  \label{fig:train_length_dis}
\end{figure}

\begin{figure}[h]
  \centering
  \includegraphics[width=.6\linewidth]{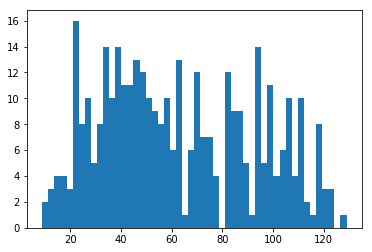}
  \caption{Distribution of sample length in test set.}
  \label{fig:test_length_dis}
\end{figure}

The most important reason we choose this dataset is the number of samples. After being preprocessed, there are a total of 1.6 million samples with a balanced distribution between positive and negative labels. This large dataset enables us to keep the experiment fair by training the model from scratch without pretraining or transfer learning. The neural network should be able to learn the hierarchy of abstract meaning for the concepts inside the sentence without any supplemental information. The test data contain 177 negative and 182 positive samples constructed manually. Figure \ref{fig:train_length_dis} and Figure \ref{fig:test_length_dis} show the distributions of the sample length of this dataset.

We preprocess the data before training. All of the nonstandard characters, such as emoji characters and hyperlinks, are removed from the text. Preprocessing is very important because many tweets contain nonstandard text, which makes the model difficult to predict. The word vectors are obtained at the word embedding layer and fed into the neural network with the architecture in Figure \ref{fig:general_architecture}. The final output is a 2-dimensional vector used for binary classification.

We train four deep learning models from scratch. The first model is a multilayer perceptron model with one hidden layer connecting the embedding layer and the output layer directly. The second model is a convolutional neural network with one convolutional layer of kernel size $100 \times 100$, one max-pooling layer with a kernel size of $20 \times 20$, and a fully connected layer. The third one is a bidirectional LSTM neural network with two stacked LSTM layers containing 128 neurons, followed by a fully connected layer. The last layer is an SCNN with two swarm filters of $300$ and $100$ neurons and a fully connected layer. 
All models are trained and validated in the same setting. The total vocabulary used in this experiment contains the 100 K most popular English words. The embedding layer converts the words to 100-dimensional vectors. The max sequence length is $140$, the batch size is $32$, the learning rate is $0.001$, and the number of epochs is 10. We use $10\%$ of the original training set as the development set, and the remaining is used to train the model.

\subsubsection{Effectiveness and Efficiency}
The first criterion to evaluate deep learning models is effectiveness. Effectiveness tells us how well the model can achieve the goal in our defined task. The more accurate the value that the model can achieve, the better the model. On the other hand, efficiency shows how optimal in cost the model should be in order to obtain the result. From our perspective, the fewer parameters the model requires to learn and perform a task, the more efficient the model. Table \ref{tab:acc_para_compare} lists the models’ number of parameters and their performance in this task.

\begin{table}
  \caption{Efficiency (number of parameters excluding embedding neurons) and effectiveness (accuracy on test set) of each model}
  \label{tab:acc_para_compare}
  \begin{center}
  \begin{tabular}{lcc}
  \hline
    \textbf{Model}& \textbf{Number of Parameters}& \textbf{Accuracy}\\
    \hline
    MLP&28,002&81.1\\
    CNN&200,142&83.3\\
    LSTM&738,307&83.6\\
    SCNN&\textbf{332}&\textbf{84.4}\\
    \hline
    \end{tabular}
   \end{center}
\end{table}

Although SCNN did not produce a large gap compared to other models in terms of effectiveness, this model has a bold difference in terms of efficiency. The SCNN achieved state-of-the-art performance and used only 332 parameters. MLP achieved 81.1\% accuracy with approximately 28 K parameters. CNN and LSTM outperformed MLP with accuracy values of 83.3\% and 83.6\% but require many more parameters (200 K and 700 K, respectively).

The swarm filter is an efficient way to reduce the dimension of the data from layer to layer. Suppose we replace swarm filters with fully connected layers. To convert the information from the embedding layer to the second layer with a dimension of 300 and the third layer with a dimension of 10, it would take 4 million parameters. Such architecture is thousands of times costlier than SCNN.

\subsubsection{Learning Curve Analysis}
\begin{figure}
\centering
  \includegraphics[width=.45\linewidth,height=100pt]{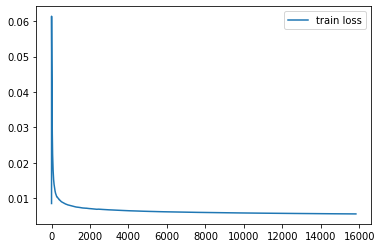}
  \includegraphics[width=.45\linewidth,height=100pt]{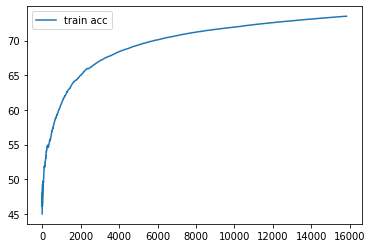}
\caption{Learning curve of MLP in first training epoch}
\label{fig:mlp_learning_curve}
\end{figure}

\begin{figure}
\centering
  \includegraphics[width=.45\linewidth,height=100pt]{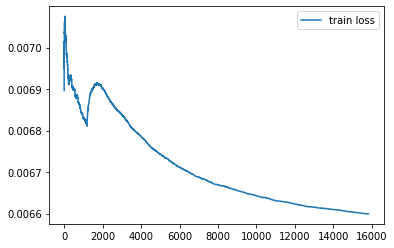}
  \includegraphics[width=.45\linewidth,height=100pt]{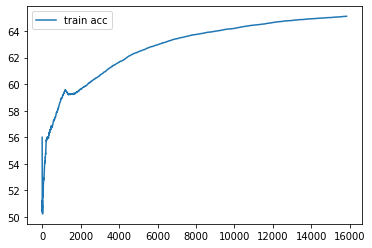}
\caption{Learning curve of CNN in first training epoch}
\label{fig:cnn_learning_curve}
\end{figure}

\begin{figure}
\centering
  \includegraphics[width=.45\linewidth,height=100pt]{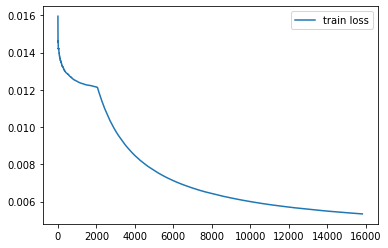}
  \includegraphics[width=.45\linewidth,height=100pt]{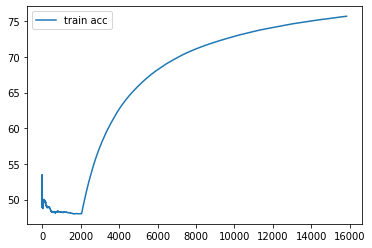}
\caption{Learning curve of LSTM in first training epoch}
\label{fig:lstm_learning_curve}
\end{figure}

\begin{figure}
\centering
  \includegraphics[width=.45\linewidth,height=100pt]{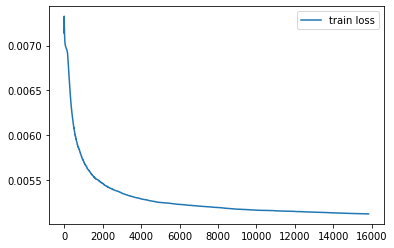}
  \includegraphics[width=.45\linewidth,height=100pt]{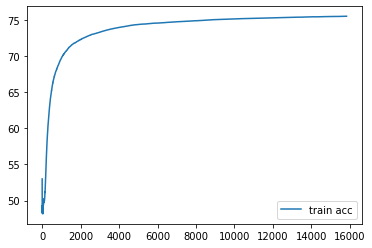}
\caption{Learning curve of SCNN in first training epoch}
\label{fig:scnn_learning_curve}
\end{figure}

We analyze the learning curve of 4 models on the training set to compare their ability to find features that represent the data. Observing the model trained with 1.6 million sentences, we found that most of the changes in accuracy and loss value occur in the first epoch. From the second epoch, these values are only adjusted by a small amount. Analyzing the learning curve in detail in the first epoch enables us to understand the learning behavior of these models as well as the level of their data hunger. Figures \ref{fig:mlp_learning_curve}-\ref{fig:scnn_learning_curve} show the learning curves of MLP, LSTM, CNN and LSTM. The accuracy and loss values were recorded after every 100 samples.

The MLP learning curve differs significantly from that of the other models. The loss increased and decreased sharply in the beginning samples. The accuracy curve in the right subfigure shows that the model then becomes stable and learns well in the subsequent sentences. The other three models had better regulation. As a result, the loss value did not swing drastically, as in MLP’s case. However, both CNN and LSTM have a struggling period before they become stable. The learning curve of the SCNN is the smoothest among all four models. This experiment reveals the learning ability of the models when compared to each other.

We made another interesting discovery when analyzing the learning curves of the models. Looking at the first 2000 iterations, we can see that the SCNN has achieved significant performance, while the rest are still struggling. After iteration 2000, SCNN performance increases but not significantly. This model is capable of identifying essential features with a few initial iterations and enhancing feature extraction if more data are available. With fewer than 2000 samples, the SCNN produces a large gap compared to the three other models.

With fewer parameters than other models, the SCNN still has good performance and efficient feature extraction and data generalization. The swarm filter was probably the origin of this result. We visualize the output of the swarm filter to better understand its behavior.

\subsubsection{Swarm Feature Visualization}
\begin{table}
  \caption{Last swarm filter output tensor}
  \label{tab:swarm_feature_examples}
  \begin{center}
  \begin{tabular}{lcc}
  \hline
    \textbf{Sentence}& \textbf{Sentiment}& \textbf{Tensor value}\\
    \hline
    I love it so much.&+& 0.14 0.15 0 0.15 0 0.14 0 0 0.21 0.15\\
    I hate it.&-&0.34 0.37 0 0.36 0 0.34 0 0 0.50 0.35\\
    I am mad at you.&-&0.25  0.27 0 0.26 0 0.25 0 0 0.37 0.26\\
    It a terrible situation.&-&0.44 0.47 0 0.47 0 0.44 0 0 0.65 0.46\\
    Today, I am very happy.&+&0.14 0.15  0 0.14 0 0.14 0 0 0.20 0.14\\
    I am sad.&-&0.45 0.48 0 0.47 0 0.45 0 0 0.66  0.46\\
    Today is a wonderful day.&+&0.11 0.12 0 0.12 0 0.11 0 0 0.16 0.12\\
    I love it.&+&0.14 0.15 0 0.15 0 0.14 0 0 0.21 0.15\\
    \hline
    \end{tabular}
   \end{center}
\end{table}

\begin{figure}[h]
  \centering
  \includegraphics[width=.6\linewidth]{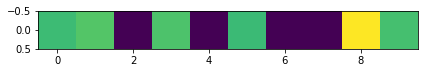}
  \caption{Plot of heatmap of last swarm features generated from positive sentence, ``I love it.” Cells with greater values are brighter.}
  \label{fig:i_love_it}
\end{figure}

\begin{figure}[h]
  \centering
  \includegraphics[width=.6\linewidth]{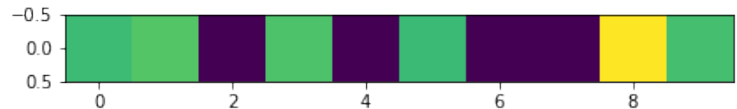}
  \caption{Plot of heatmap of last swarm features generated from negative sentence, ``I am sad.” Cells with greater values are brighter.}
  \label{fig:i_am_sad}
\end{figure}

Calculating the features using Equation \ref{eq:scnn}, the swarm filter makes them follow the ratio of nodes in the filter as a simple pattern. For a better illustration of the generalization capability of SCNN, we visualize the features generated by swarm filters. Table \ref{tab:swarm_feature_examples} lists some example inputs and their sentiment class and the corresponding tensors outputted by the last swarm filter. All tensors are parallel to each other regardless of whether the data are positive or negative. Figure \ref{fig:i_love_it} and Figure \ref{fig:i_am_sad} show heatmaps of the tensors in the positive and negative cases; there is no difference between them.

The visualization experiment shows the ability of the SCNN to abstract the characteristic features of individuals belonging to a latent swarm. Passing the data through two swarm filters, the model views every sample from the same aspect. In other words, it simplifies the pattern of the data. The last swarm filter’s output tensor is passed into a dense layer to obtain the final decision. The SCNN collects and abstracts the features of individuals effectively before making the final prediction. Such an approach could be the reason that the SCNN requires fewer parameters but performs better than other architectures.

\section{RELATED WORKS}
\label{sec:related_works}

Compared to the ideas of SCNN in this paper, previous works have applied the idea of swarm intelligence to neural networks in different ways. Swarm optimization algorithms can help to find optimized parameters for deep learning. Singh et al. \cite{singh2018integration}, Ye et al. \cite{ye2017particle} and Kenny et al. \cite{kenny2017study} are some of the authors who made proposals using this approach.

Singh et al. proposed the follow-the-leader (FTL) algorithm to improve the effectiveness of the artificial neural network in forecast problems. The algorithm is inspired by the movement of sheep belonging to a herd. Each sheep should follow his leader, which is the sheep with the best moves. In detail, after the feedforward neural network calculates the error, the configuration is then optimized by FTL.

With the same purpose of optimizing the structure of deep neural networks, Ye et al. proposed a method using the PSO algorithm. In their proposal, each particle independently represents a configuration of the neural networks. Hyperparameters of the model are set to the optimal value by the algorithm.

In another work, autoencoder networks were enhanced by Kenny et al. using the PSO approach. They conducted a study investigating the effects of the method in a high-dimensional space. The network was divided into smaller subnetworks with a lower number of connection weights. After that, subnetworks were optimized using PSO.

Swarming is limited not only in swarm intelligence and the dynamic aspect but also in the swarm feature and the static aspect. Such information is not only surface information but also latent features that require extraction to be understood. From the above observation, the swarm filter \cite{nguyen2019swarm} was proposed as an innovative way of combining swarming and deep learning research.

\section{CONCLUSION}
\label{sec:conclusion}
By conducting detailed experiments, we verified the effectiveness and efficiency of the SCNN. This architecture outperformed models using standard deep learning components such as the multilayer perceptron, LSTM, and CNN. Furthermore, we showed that the SCNN requires fewer parameters and fewer training data.

There are some future directions for the SCNN and swarm filter. With its excellent ability to extract features, this architecture could contribute to more elaborate designs such as transformer-based models \cite{vaswani2017attention} for performance improvements. The lightweight property of this model can benefit a system with limited storage. In addition, the learning ability of this model is useful for domains in which training data are scarce or costly to obtain.

\bibliographystyle{plain}      
\bibliography{./references}   

\end{document}